\documentclass[conference]{IEEEtran}
\IEEEoverridecommandlockouts

\usepackage{graphicx}
\usepackage{booktabs}
\usepackage{amssymb}
\usepackage{makecell}
\usepackage{multirow}
\usepackage{array}
\usepackage{pgfplots}
\usepackage{amsmath}
\def\BibTeX{{\rm B\kern-.05em{\sc i\kern-.025em b}\kern-.08em
    T\kern-.1667em\lower.7ex\hbox{E}\kern-.125emX}}
\begin{document}
\graphicspath{{Figures/}}

\title{Categorical Knowledge Fused Recognition: Fusing Hierarchical Knowledge with Image Classification through Aligning and Deep Metric Learning}

\author{\IEEEauthorblockN{Yunfeng Zhao}
\IEEEauthorblockA{
\textit{Zhejiang University}\\
zyfccc@126.com}
\and
\IEEEauthorblockN{Huiyu Zhou}
\IEEEauthorblockA{
\textit{University of Leicester}\\
hz143@leicester.ac.uk}
\and
\IEEEauthorblockN{Fei Wu}
\IEEEauthorblockA{
\textit{Zhejiang University}\\
wufei@cs.zju.edu.cn}
\and
\IEEEauthorblockN{Xifeng Wu}
\IEEEauthorblockA{
\textit{Zhejiang University}\\
xifengw@zju.edu.cn}
}

\maketitle

\begin{abstract}
    Image classification is a fundamental computer vision task and an important baseline for deep metric learning. In decades efforts have been made on enhancing image classification accuracy by using deep learning models while less attention has been paid on the reasoning aspect of the recognition, i.e., predictions could be made because of background or other surrounding objects rather than the target object. Hierarchical knowledge about image categories depicts inter-class similarities or dissimilarities. Effective fusion of such knowledge with deep learning image classification models is promising in improving target object identification and enhancing the reasoning aspect of the recognition. In this paper, we propose a novel deep metric learning based method to effectively fuse prior knowledge about image categories with mainstream backbone image classification models and enhance the reasoning aspect of the recognition in an end-to-end manner. Existing deep metric learning incorporated image classification methods mainly focus on whether sampled images are from the same class. A new triplet loss function term that aligns distances in the model latent space with those in knowledge space is presented and incorporated in the proposed method to facilitate the dual-modality fusion. Extensive experiments on the CIFAR-10, CIFAR-100, Mini-ImageNet, and ImageNet-1K datasets evaluated the proposed method, and results indicate that the proposed method is effective in enhancing the reasoning aspect of image recognition in terms of weakly-supervised object localization performance.
\end{abstract}


\section{Introduction}

   \begin{figure}[t]
      \centering
      \includegraphics[width=0.95\linewidth]{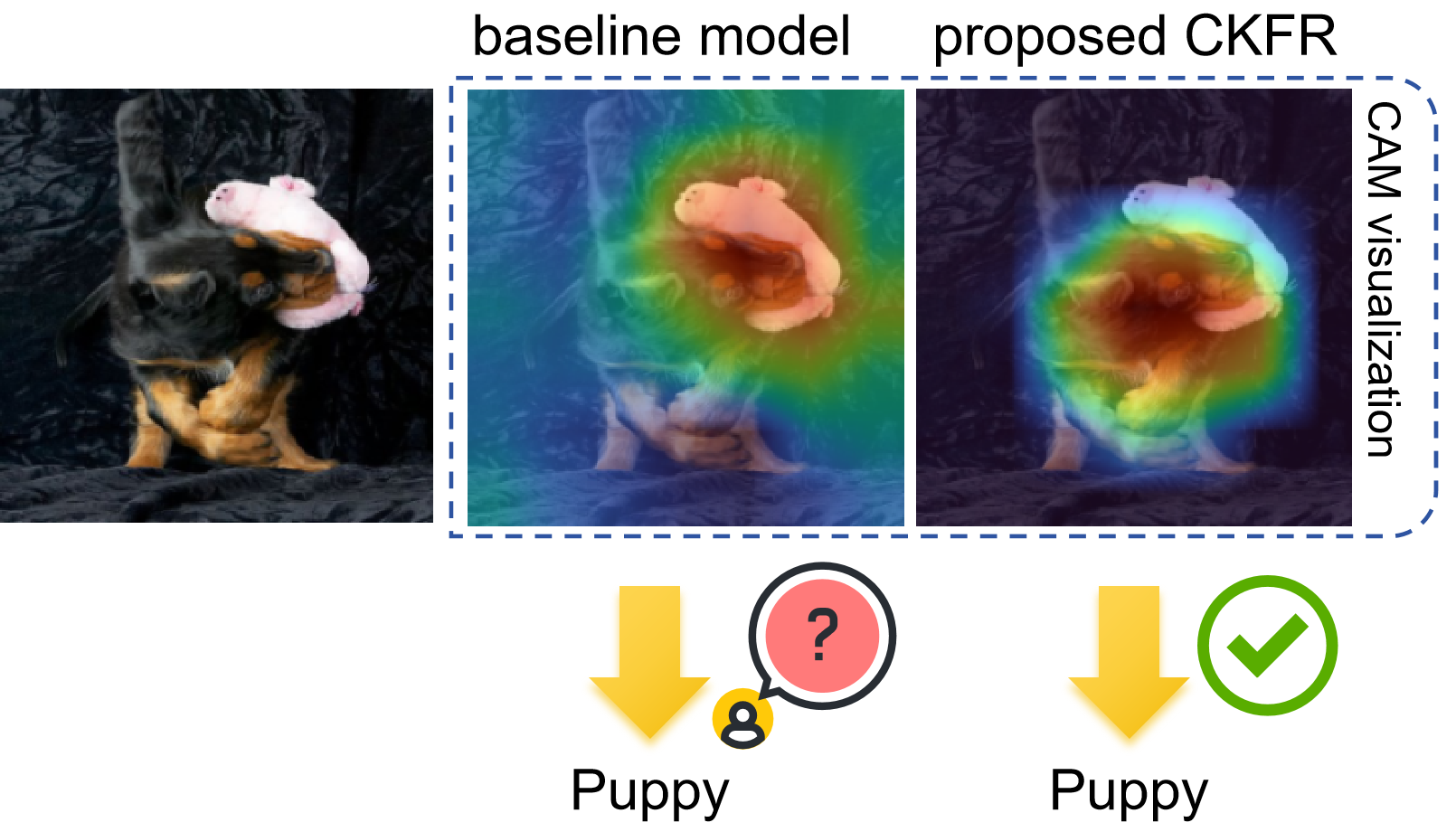}
      \caption{Illustration of the proposed method that identifies the target object rather than background or surrounding objects in the recognized image for enhancing the reasoning aspect of image recognition and improving prediction robustness.}
      \label{fig:intro}
   \end{figure}

   Image classification attempts to correctly predict the belonging category of images \cite{2015ImageNet}. Image classification based on deep learning has achieved remarkable performance by a range of deep learning based backbone models such as Convolutional Neural Networks \cite{ResNet2016,Alexnet-2012,1989Backpropagation} and Vision Transformers \cite{cvt2021,VTrans2021}. The training process of these image classification models is usually supervised.

   Knowledge about image categories such as category trees and semantic hierarchies captures inter-class similarities or dissimilarities. It is a potentially valuable source of context information that can improve recognition performance \cite{8825856}. The major current approach to conduct such incorporation is through constructing a corresponding tree-shaped hierarchical classifiers that performs multiple classification steps to make the prediction \cite{8825856,Qu-2017,Sun-2013}. Effective fusion of such knowledge with image classification in an end-to-end manner remains a challenge.

   Image classification focusing on improving classification accuracy only can make prediction based on features in the image that is correlated to the category to be predicted other than the target object \cite{causal-att-2021}. For example, a street road background is very likely to associate with cars while this is not a causal clue for classifying an image to the car category. Such confounding factors can mislead the predicting model especially in out-of-distribution real-world applications, and causal image classification is often facilitated by conducting dataset splitting or masking to mitigate confounders or emphasize the target object \cite{causal-survey}. Weakly-supervised object localization (WSOL) aims to localize objects in images mostly by using image-level priors \cite{PSOL-2020,fam-2021} and can be a task that facilitates the reasoning aspect of image recognition. Figure \ref{fig:intro} demonstrates an example and effects of the proposed method on improving the reasoning aspect of image recognition.

   The sense of distance has been adopted in image classification. Deep contrastive and metric learning alter model latent space. They have been applied on deep image classification backbone models to improve recognition performance \cite{c2am-2022,Seo-2022,cl-2021,DetCo-2021}. Existing loss function terms in deep contrastive and metric learning such as contrastive and triplet losses mainly utilize discriminativity, i.e., pulling closer deep learning model latent representations of similar samples and pushing apart those of dissimilar ones \cite{c2am-2022, dynamic-2022, Liu-2019, 9405473, sym11091066}, to embrace such distance measurements in the deep models. However, rich similarity and dissimilarity information in knowledge about image categories is lost when these current mechanisms are applied. 
  
   In this paper, we propose a novel method, namely, Categorical Knowledge Fused Recognition (CKFR), that fuses knowledge about image categories with deep image classification backbone models through aligning and deep metric learning. Aligning is a weakly-supervised deep metric learning process that achieves the relative similarity between the model latent and knowledge spaces. Compared to the major existing image classification approaches, the proposed method can effectively fuse knowledge about image categories with deep image classification models in an end-to-end manner and enhance the reasoning aspect of the recognition, i.e., it produces image classification models that more accurately localize the target objects in the recognized images based on class activation map (CAM) \cite{cam-2016}. The major contributions of this work are summarized below:
  
   \begin{enumerate}
      \item Effective fusion of inter-class similarity prior knowledge with backbone image classification model is achieved through aligning and deep metric learning in an end-to-end manner. The fusion is achieved by learning a model latent space whose inter-class similarities are aligned with those in the knowledge space.
      \item A new triplet loss function term that aligns distances in the model latent space with those in the inter-class similarity knowledge space is proposed. The proposed triplet loss term prevents any additional variable, e.g., a margin term.
      \item Extensive experiments that evaluate the proposed method have been conducted on CIFAR-10, CIFAR-100, Mini-ImageNet, and ImageNet-1K datasets. The proposed method provoked 11.0\% and 14.4\% improvements on Top-1 and Top-5 localization accuracies compared to the baseline, indicating its effectiveness on enhancing the reasoning aspect of image recognition.
   \end{enumerate}
   
   \section{Related Works}
   \label{sec:related-works}
   
   \subsection{Metric learning}
   Manhattan and Euclidean, special cases of Minkwoski and also known as $L_1$ and $L_2$ distance, are basic distance metrics \cite{metric-learning}. Common distance metrics in metric learning include cosine similarity \cite{metric-learning}, Mahalanobis \cite{metric-learning-survey}, Kullback-Leibler Divergence \cite{KLD2022}, etc.
   
   Existing supervised deep metric learning based methods applied for image recognition tasks have been mainly exploiting discriminativity \cite{dynamic-2022,Liu-2019,features-2016} (or qualitative relativity \cite{metric-learning,schultz-2003}) in image feature \cite{quadruplets-2017} or class \cite{dynamic-2022,features-2016,metric-learning-survey, schultz-2003} level. Contrastive \cite{dynamic-2022}, triplet \cite{Aurora2022,schultz-2003}, or quadruplet \cite{quadruplets-2017} losses have been incorporated in these deep metric learning methods to rearrange the model latent space.
   
   Several works also employed contrastive loss with self-supervised \cite{Wu2022,DetCo-2021,DiRA-2022} or unsupervised learning \cite{c2am-2022,MoCo-2020,unsupervised-2018} to improve model latent representation and performance of visual tasks. For instance, Xie et al. \cite{c2am-2022} proposed an object localization method by exploiting cross-image foreground-background contrast and employed a contrastive loss to learn disentangled representations of foreground objects \cite{c2am-2022}.
   
   \subsection{Contextual knowledge incorporation}
   Contextual knowledge has various types and forms and can be fused with image classification in multiple ways. Objects and the surrounding environment in the image to be classified can be identified to improve classification performance \cite{word-text-2017,semantic-attri-2012}. Graph embedding of objects/features in the image to be classified can also improve image classification performance \cite{graph-emb-2021,graph-2021}. Another way of integrating contextual knowledge with image classification is through dense embedding and fusion of textual and semantic knowledge about image features in language models \cite{image-text-2024,CLIP-FG-2023,CMA-CLIP-2022,image-text-2019}. Hierarchical knowledge such as semantic hierarchy captures inter-class similarities knowledge about image categories and is valuable to many visual tasks \cite{dynamic-2022,Qu-2017}. 
   
   To incorporate hierarchical knowledge and enhance image recognition performance, multiple works applied various mechanisms. Deng et al. \cite{semantic-indexing-2011} defined the hierarchical precision metric and utilized the form of bilinear similarity to learn a correspondence between latent representations and semantic hierarchy through metric learning. Waltner et al. \cite{HiBsteR-2019} incorporated hierarchical semantic knowledge with backbone visual model by adapting boosting mechanism, where coarse and fine embeddings were separately defined in layers and their weights updated by applying deep metric learning. Other works constructed classifiers with the corresponding tree structure that require multiple steps to perform a prediction \cite{8825856,hyperspectral2019,Qu-2017,Sun-2013}.
   
   \subsection{Knowledge fusion}
   Data fusion attempts to create proper interactions, either through soft or hard links, between datasets with various modalities \cite{data-fusion-2015}. Few works fused hierarchical knowledge about image categories with backbone visual models by optimizing model latent space. Zheng et al. \cite{dynamic-2022} proposed a hierarchical concept refiner mechanism that distills latent embedding to hierarchical layers of concepts according to hierarchical semantic knowledge. Kim et al. \cite{HIER-2023} mapped model latent space to a low-dimensional hyperbolic space and utilized hyperbolic distance metric and deep metric learning to alter the backbone model latent representation to a semantic hierarchy structure. This method calculates the lowest common ancestors of pairing samples as proxies in addition to the sampled triplets. 
   
   The proposed method in this paper fuses inter-class similarity knowledge with backbone deep learning image classification models through aligning the model latent space with knowledge space by utilizing deep metric learning. It can be performed directly on high-dimensional model latent representation, with arbitrary triplet sampling, and in an end-to-end manner. In contrast with most existing deep metric learning based methods that optimize latent space of backbone image classification models by utilizing discriminativity in image classes, the proposed method numerically aligns model latent space with the inter-class similarity knowledge space through deep metric learning.

   \section{Proposed Method}
   \label{sec:methods}
  
   In this section, the proposed CKFR method that fuses inter-class similarity knowledge about image categories with deep learning based image classification backbone models is presented. In general, the proposed CKFR method learns a model latent space whose inter-class similarities are aligned with those in the knowledge embedding space through deep metric learning and in an end-to-end manner.

   The latent spaces of deep learning backbone models are usually encoded with high dimensionality and often normalized, making the distance measurement in the model latent space varied from that in knowledge space. Therefore, distance metrics of latent representations and knowledge embeddings are firstly defined to enable distance measurement in these two spaces. A new triplet loss term is then introduced to align these two spaces.  

   \subsection{Preliminaries}

   As a proof of concept, the prior knowledge embedding space of inter-class similarities has the form of a tree structure $\mathcal{T} =\left(O, E\right)$, where $o\in O$ is a node/leaf denoting an image category and $e\in E$ is a weighted edge denoting pseudo-distance between a node/leaf and its parent node. The knowledge tree $\mathcal{T}$ incorporates hierarchical prior knowledge about image inter-class similarities. And it can be processed to produce a symmetric pseudo-distance matrix with zero-diagonal $D^{\mathcal{T}}$, or a prior matrix, that depicts inter-class distances.

   Assume an image dataset $\mathcal{X} = \left\{\left(\boldsymbol{X}_i, \boldsymbol{y}_i\right)\right\}^n_i$ of $n$ samples with corresponding image class label $\boldsymbol{y}, \forall \boldsymbol{y} \in O$, an image classification backbone model for the image classification task obtains the model latent vector in the latent space $\boldsymbol{z} \in \mathbb{R}^m$ by an image feature extractor $Extract\left(\boldsymbol{X}\right)$, where $m$ is the dimensionality of the model latent space. The latent representation is connected to a classifier, e.g., $MLP\left(\boldsymbol{z}\right)$, to produce prediction results.

   \subsection{Distance metrics}
   \label{sec:dis-metric}
   
   Two distance metrics are defined in this work. The first metric measures distance in the knowledge embedding space, and the other quantifies distance in the model latent space.
   
   In this work, distance between two categories $o^a$ and $o^b$ in the knowledge embedding space is measured by a pseudo-distance metric. The metric measures the shortest edge distance traveled between two corresponding tree nodes/leafs of the categories, denoted by:
   \begin{equation}\label{equ:knowledge}
      d^\mathcal{T}\left(o^a, o^b\right) = \sum \boldsymbol{e}, \forall \boldsymbol{e} \in E
   \end{equation}
   where $\boldsymbol{e}$ is the edge vector containing shortest edges need to be traveled between two measured nodes.
   
   Since the distances in the prior matrix will be nonlinearly transformed by a deep neural network, a constant pseudo-distance value $k$ can be assigned to all the weighted edges in the knowledge tree. 
   
   The latent representation of a deep learning model is high in dimensionality. The latent distance metric is defined to measure distance in the model latent space. In the defined latent distance metric, the model latent vector $\boldsymbol{z}$ of a batch is initially normalized by its mean $\mu$ and standard derivation $\sigma$. The normalization is denoted in Equation \ref{equ:normalization}:
   \begin{equation}\label{equ:normalization}
      \begin{gathered}
         \mu = \frac{\sum_{i=1}^{m} \boldsymbol{z}_i}{m}\\
         \sigma = \sqrt{\frac{\sum_{i=1}^{m}\left(\boldsymbol{z}_i - \mu\right)^2}{m}}\\
         \boldsymbol{\Hat{z}} = \frac{\left(\boldsymbol{z} - \mu \right)}{\sigma } 
      \end{gathered}
   \end{equation} 
   where $\boldsymbol{\Hat{z}}$ is the normalized latent vector.
   
   Then, the distance between two normalized latent vectors $\boldsymbol{\hat{z}}^a$ and $\boldsymbol{\hat{z}}^b$ in the model latent space has the form:
   \begin{gather}
      d^z\left(\boldsymbol{z}^a, \boldsymbol{z}^b\right) = \| \boldsymbol{\Hat{z}}^a - \boldsymbol{\Hat{z}}^b\|
   \end{gather}
   where $\|$ denotes distance measurement and can be calculated as:

   \begin{gather}
      \| \boldsymbol{\Hat{z}}^a - \boldsymbol{\Hat{z}}^b\| = \sum_{i=1}^{m} \left(|\boldsymbol{\Hat{z}}^a_i - \boldsymbol{\Hat{z}}^b_i | \right)^\ell
   \end{gather}
   where $|$ denotes taking positive values, $\ell$ is the exponent, and $\ell \geq1$.

   The defined metrics satisfy the nonnegativity, identity of indiscernibles, symmetry, and triangle inequality distance axioms.

   \subsection{Quantitative-relativity triplet loss incorporated deep metric learning}
   
   \begin{figure*}[htbp]
      \centering
      \includegraphics[width=0.82\linewidth]{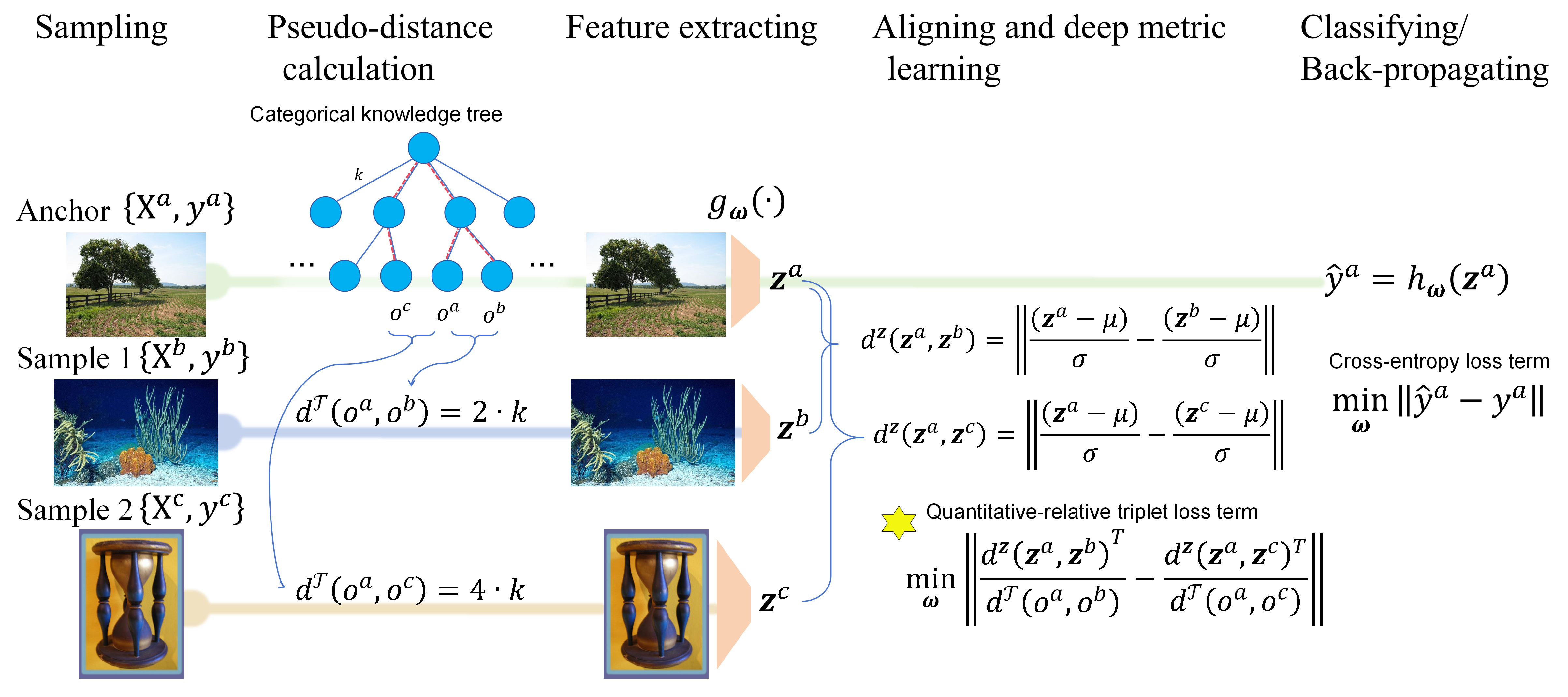}
      \caption{The training process of the proposed method is illustrated by using a single triplet as an example. The initial step samples the triplet. The following step calculates pseudo-distances between sample categories in the knowledge tree. In the third step, images are encoded to latent representations by the feature extractor. In the next step, distances between sample latent representations are measured, and the proposed quantitative-relativity triplet loss is applied to align the knowledge space with the model latent space by deep metric learning. A cross-entropy loss is also applied for supervised image classification learning. }
      \label{fig:pipeline}
   \end{figure*}
   
   Figure \ref{fig:pipeline} demonstrates the overall training process of the proposed CKFR method. The method fuses inter-class similarity knowledge about image categories with deep image classification models by aligning distances in knowledge space with those in model latent space. 
   
   In the training phase, the proposed method randomly samples an anchor and two pairing samples from the dataset $\mathcal{X}$ to form the triplet $\mathcal{P} = \left\{(\boldsymbol{X}^a_i, \boldsymbol{y}^a_i), (\boldsymbol{X}^b_i, \boldsymbol{y}^b_i), (\boldsymbol{X}^c_i, \boldsymbol{y}^c_i)\right\}^n_i$, where $(\boldsymbol{X}^a, \boldsymbol{y}^a)$ denotes the anchor, $(\boldsymbol{X}^b, \boldsymbol{y}^b)$ and $(\boldsymbol{X}^c, \boldsymbol{y}^c)$ refer to the other two pairing samples.
   
   In the proposed method, feature extractor with nonlinear activation is applied to enable nonlinear mapping $g$ from the image data to the model latent space. The process is denoted by:
   \begin{equation}
      \boldsymbol{z}=g_{\boldsymbol{\omega}}\left(\boldsymbol{X}\right), \boldsymbol{z} \in \mathbb{R}^m
   \end{equation}
   
   A classifier $h$ is integrated to make the classification predictions:
   
   \begin{equation}
      \boldsymbol{\hat{y}} = h_{\boldsymbol{\omega}}\left(\boldsymbol{z}\right)
   \end{equation}
   
   The proposed CKFR method constrains the model weights $\bold{\omega}$ of the nonlinear activated feature extractor and produces a model latent space that satisfies the alignment criteria in terms of the model latent and pseudo-distance metrics defined in Section \ref{sec:dis-metric}. The alignment is scalable by a scaling factor, and the criteria can be denoted as:
   \begin{equation}\label{equ:alignment}
      d^z\left(\boldsymbol{z}^a, \boldsymbol{z}^b\right) = \lambda d^\mathcal{T}\left(o^a, o^b\right)
   \end{equation}
   where $\lambda$ is the scaling factor.
   
   The alignment criteria in Equation \ref{equ:alignment} can be further derived to formulate Equation \ref{equ:qtr-modeling} by considering the other pairing sample in the triplet:
   
   \begin{equation}\label{equ:qtr-modeling}
      \frac{d^z\left(\boldsymbol{z}^a, \boldsymbol{z}^b\right)^T}{d^\mathcal{T}\left(o^a, o^b\right)} = \frac{d^z\left(\boldsymbol{z}^a, \boldsymbol{z}^c\right)^T}{d^\mathcal{T}\left(o^a, o^c\right)}
   \end{equation}
   
   Based on this setting, a new quantitative-relativity loss $\mathcal{L}_{qtr}$ with the triplet form is proposed to fuse prior knowledge about image categories with deep image classification model weights. The quantitative-relativity triplet loss can be formulated as:
   
   \begin{gather}
      \mathcal{L}{qtr}\left(\boldsymbol{z}^a, \boldsymbol{z}^b, \boldsymbol{z}^c\right) = \mathcal{L}{qtr}\left(\boldsymbol{z}^a, \boldsymbol{z}^c, \boldsymbol{z}^b\right)\nonumber \\
         = \| d^z\left(\boldsymbol{z}^a, \boldsymbol{z}^b\right)^T d^\mathcal{T}\left(o^a, o^c\right) - d^z\left(\boldsymbol{z}^a, \boldsymbol{z}^c\right)^T d^\mathcal{T}\left(o^a, o^b\right) \| 
   \end{gather}
   where $\boldsymbol{z}^a$ is the latent vector of the anchor, and $L_1$ distance can be used to calculate the loss. Switching positions of $\boldsymbol{z}^b$ and $\boldsymbol{z}^c$ will produce identical result. All the triplet inputs can be sampled from arbitrary image categories. 
   
   The proposed quantitative-relativity triplet loss term encourages an alignment between inter-class similarities in the latent space and those in the prior knowledge space. Therefore, a merit of the proposed loss is that it prevents any additional absolute variable, for example, a margin term that needs to be tuned according to specific learning task and could bring additional bias to the system, when performing the alignment.
   
   The objective function of the learning can be written as:
   
   \begin{equation}
      \label{equ:obj-func}
      \min_{\boldsymbol{\omega}} \left(\mathcal{L}_{ce}\left(\boldsymbol{\hat{y}}^a, \boldsymbol{y}^a\right) + \alpha \cdot \left( \mathcal{L}_{qtr}\left(\boldsymbol{z}^a, \boldsymbol{z}^b, \boldsymbol{z}^c\right)\right) \right)
   \end{equation}
   where $\mathcal{L}_{ce}$ denotes a cross-entropy loss term between the predicted and corresponding label values, $\alpha$ denotes the weighting hyperparameter of the quantitative-relativity loss term.
   
   Note that the scalar part of the constant pseudo-distance value $k$ defined in Section \ref{sec:dis-metric} can be extracted and combined with the weighting hyperparameter $\alpha$.
   
   The objective function defined in Equation \ref{equ:obj-func} can be extended to contain more than one cross-entropy and quantitative-relativity loss term by switching between anchor and pairing samples.

   \section{Experiments}
   \label{sec:experiments}
   
   In this section, classification and weakly-supervised object localization performances of the proposed method were evaluated by numeral image datasets and compared with baselines based on major existing image classification backbone models.
   
   \subsection{Datasets}
   \label{sec:datasets}
   
   Four image datasets with general image categories, i.e., CIFAR-10 \cite{CIFAR}, CIFAR-100 \cite{CIFAR}, Mini-ImageNet \cite{Mini-imagenet}, and ImageNet-1K \cite{2015ImageNet} were used to evaluate the proposed CKFR method. The CIFAR-10 dataset consists of 10 general image categories with $6,000$ images of $32\times32$ in size. The CIFAR-100 dataset contains a hundred image categories with each one containing $600$ images of the same size. The number of image categories scales up to a thousand in the ImageNet-1K dataset, with each one containing approximately $1,300$ images of around 300 pixels in the shorter edge. The Mini-ImageNet dataset is a subset of ImageNet-1K and also contains a hundred image categories. The number of images per category in the Mini-ImageNet dataset is around $1,800$, and the images are with similar sizes as those in the ImageNet-1K. 
   
   The hierarchical knowledge tree of image categories used in this work was produced by ImageNet \cite{2015ImageNet} based on WordNet synsets \cite{miller-1995}. The object bounding boxes within the ImageNet-1K dataset were used as the ground truth bounding boxes of the target objects for the WSOL performance evaluation.

   \subsection{Dataset preparation and preprocessing}
   
   In this work, image categories in the CIFAR-10 and CIFAR-100 datasets were manually mapped with the closest corresponding categories in the semantic knowledge tree to enable the conducting of the evaluations on the CIFAR-10 and CIFAR-100 datasets. A constant $1.0$ was assigned to the pseudo-distance $k$ in the knowledge tree. A pseudo-distance matrix $D^T$ was established based on the knowledge tree to accelerate training process.
   
   Training and validation splitting remained original as it was in CIFAR-10, CIFAR-100, and ImageNet-1K datasets. In the Mini-ImageNet dataset, the splitting was performed according to the 80/20 percentage ratio.

   \subsection{Backbone models}
   
   The backbone model of image feature extractor deployed in experiments were Convolutional Vision Transformer (CvT) \cite{cvt2021}, ResNet \cite{ResNet2016}, Wide ResNet (WResNet) \cite{wideResNet}, and Vision Transformer (ViT) \cite{VTrans2021}. Random dropout \cite{srivastava-2014} probability of $15\%$ was applied to CvTs during training to improve model generosity. ResNet-34 was selected as the backbone model for the evaluations on the CIFAR-100 dataset. WResNet-50 model was selected and fine-tuned on its pretrained weights to evaluate the proposed method on the Mini-ImageNet dataset. WResNet-101, ViT-B/16, and their pretrained weights were used to evaluate the proposed method on the ImageNet-1K dataset.

   \subsection{Training settings}
   
   The CvT and ResNet-34 model weights were trained from scratch on CIFAR datasets. They were trained with an initial learning rate of 1E-3 and an ending learning rate of 1E-4. The WResNet model weights pretrained on ImageNet were fine-tuned by a learning rate of 1E-5. 
   
   Random sampling of the triplet in the proposed CKFR method was performed with equal probability. The hyperparameter $\alpha$ and $\ell$ were selected according to the image dataset and backbone model used.
   
   The input images were resized to $64\times 64$ for CIFAR datasets and $224\times 224$ for Mini-ImageNet and ImageNet-1K datasets. Standard data augmentation operations such as random resized cropping, random horizontal flipping, random autocontrast, random rotation, and random Gaussian blurring were applied to the training image data. The CvT, ResNet-34, and WResNet-50 models were trained with Sharpness-Aware Minimization (SAM) \cite{2021Sharpness} to perform distribution smoothing and enhance model generosity. A batch size of 32 was used for all the datasets.
   
   \subsubsection{Evaluation metrics}
   
   Top-1 and Top-5 classification accuracy (Acc.) produced in validation was considered as the evaluation metric of classification performance. 
   
   $MaxBoxAcc$ \cite{WSOL2023} was selected as the evaluation metric to measure object localization accuracy. The metric is formulated as:
   
   \begin{equation}
      MaxBoxAcc(\tau, \delta) = 100\% \times \frac{1}{n} \sum_{i=1}^{n}1_{IoU(\hat{B}(\boldsymbol{x}_i,\tau),B^i)\geqslant \delta }
   \end{equation}
   where $\hat{B}(\boldsymbol{x}_i,\tau)$ are the estimated bounding boxes with activation map threshold $\tau$, $B$ denotes ground truth bounding boxes of target objects in each recognized image, $\delta$ is the intersection proportion threshold between $\hat{B}$ and $B$.

   To better identify between target object and background activations, two groups of thresholds were set to produce two levels of estimated bounding boxes, and their intersection proportion was also measured and thresholded by 0.4 to effectively include only the valid class activations on the target object rather than on the surrounding background. For WResNet model, the thresholds $\tau_{g1}$ and $\tau_{g2}$ were set to be 0.4 and 0.6, and $\delta$ was set to be 0.25. For ViT model, the thresholds $\tau_{g1}$ and $\tau_{g2}$ were set to be 0.6 and 0.7, and $\delta$ was set to be 0.15.

   GT-loc was used to evaluate object localization performance according to the ground truth object bounding boxes in the validation set. Top-1 and Top-5 localization accuracies (Top-1 and Top-5 Loc. Acc.), calculated as the percentage of images correctly classified in terms of Top-1 and Top-5 Acc. and the target object correctly localized in terms of GT-loc, were used to measure the overall object localization performance.

   \subsection{Experimental results}
   
   \subsubsection{Effectiveness evaluations of the proposed method}
   
   \newcolumntype{?}{!{\vrule width 1pt}}
   \begin{table*}[htbp]
      \centering
      \scalebox{0.9}{
      \begin{tabular}{m{2.6cm}|m{1.0cm}|m{1.0cm}?m{2.0cm}|m{2.4cm}|m{1.5cm}|m{1.5cm}}
      \hline
      \centering\multirow{2}{*}{\large Backbone} & \centering\multirow{2}{*}{\normalsize Params} & \centering\multirow{2}{*}{\normalsize FLOPs}& \centering\multirow{2}{*}{\large Dataset} & \centering\multirow{2}{*}{\large Criteria} & \multicolumn{2}{c}{\large Method} \\
      \cline{6-7}
       & & & & & \makecell[c]{ Baseline } & \makecell[c]{ CKFR } \\
      \hline\hline
      \multirow{2}{*}{CvT} & \multirow{2}{*}{$0.8$M} & \multirow{2}{*}{$81.3$M} & \centering\multirow{2}{*}{CIFAR-10} & \centering Top-1 Acc. & \centering\arraybackslash $84.8$ & \centering\arraybackslash $\bold{86.6}$ \\
       &  &  &  & \centering Top-5 Acc. & \centering\arraybackslash $99.2$ & \centering\arraybackslash $\bold{99.5}$ \\
       \hline
       \multirow{2}{*}{ResNet-34} & \multirow{2}{*}{$21.3$M} & \multirow{2}{*}{$300.3$M} & \centering\multirow{2}{*}{CIFAR-100} & \centering Top-1 Acc. & \centering\arraybackslash $58.3$ & \centering\arraybackslash $\bold{59.6}$ \\
       &  &  &  & \centering Top-5 Acc. & \centering\arraybackslash $82.8$ & \centering\arraybackslash $\bold{83.2}$ \\
      \hline
      \multirow{3}{*}{WResNet-101*} & \multirow{3}{*}{$126.9$M} & \multirow{3}{*}{$22.8$B} & \centering\multirow{3}{*}{ImageNet-1K} & \centering GT-loc & \centering\arraybackslash $56.8$ & \centering\arraybackslash $\bold{72.6}$\\
       &  &  &  & \centering Top-1 Loc. Acc. & \centering\arraybackslash $50.1$ & \centering\arraybackslash $\bold{61.1}$\\
       &  &  &  & \centering Top-5 Loc. Acc. & \centering\arraybackslash $55.5$ & \centering\arraybackslash $\bold{69.9}$\\
       \hline
       \multirow{3}{*}{ViT-B/16*} & \multirow{3}{*}{$58.1$M} & \multirow{3}{*}{$11.3$B} & \centering\multirow{3}{*}{ImageNet-1K} & \centering GT-loc & \centering\arraybackslash $28.1$ & \centering\arraybackslash $\bold{38.8}$\\
       &  &  &  & \centering Top-1 Loc. Acc. & \centering\arraybackslash $21.4$ & \centering\arraybackslash $\bold{28.8}$\\
       &  &  &  & \centering Top-5 Loc. Acc. & \centering\arraybackslash $25.8$ & \centering\arraybackslash $\bold{35.3}$\\
       \hline
       \multicolumn{6}{l}{\makecell[l]{$*$ indicates backbone model weights pretrained on ImageNet. }}
      \end{tabular}
       }
      \vspace{5pt}
      \caption{Classification and localization accuracies produced by baseline and the proposed method compared.}
      \label{tab:first-exp}
   \end{table*}
  
   The effectiveness of the proposed method was evaluated by comparing the classification and localization accuracies of the baselines and corresponding models trained by applying the proposed method. The evaluations were performed on the CIFAR-10, CIFAR-100, and ImageNet-1K datasets and their results are indicated in Table \ref{tab:first-exp}. The proposed CKFR method improved image classification accuracy on the CIFAR datasets. And evaluation results on the ImageNet-1K dataset indicate that the proposed method tremendously enhanced WSOL performance in terms of GT-loc which improved from 56.8\% to 72.6\% and 28.1\% to 38.8\% for the WResNet and ViT models, respectively. Improvements of 11.0\% and 14.4\% on Top-1 and Top-5 Loc. Acc. were achieved by using WResNet, and those of 7.4\% and 9.5\% were achieved by using ViT.

   \subsubsection{Latent space visualization}

   \begin{figure}[htbp]
      \centering
      \includegraphics[width=1.0\linewidth]{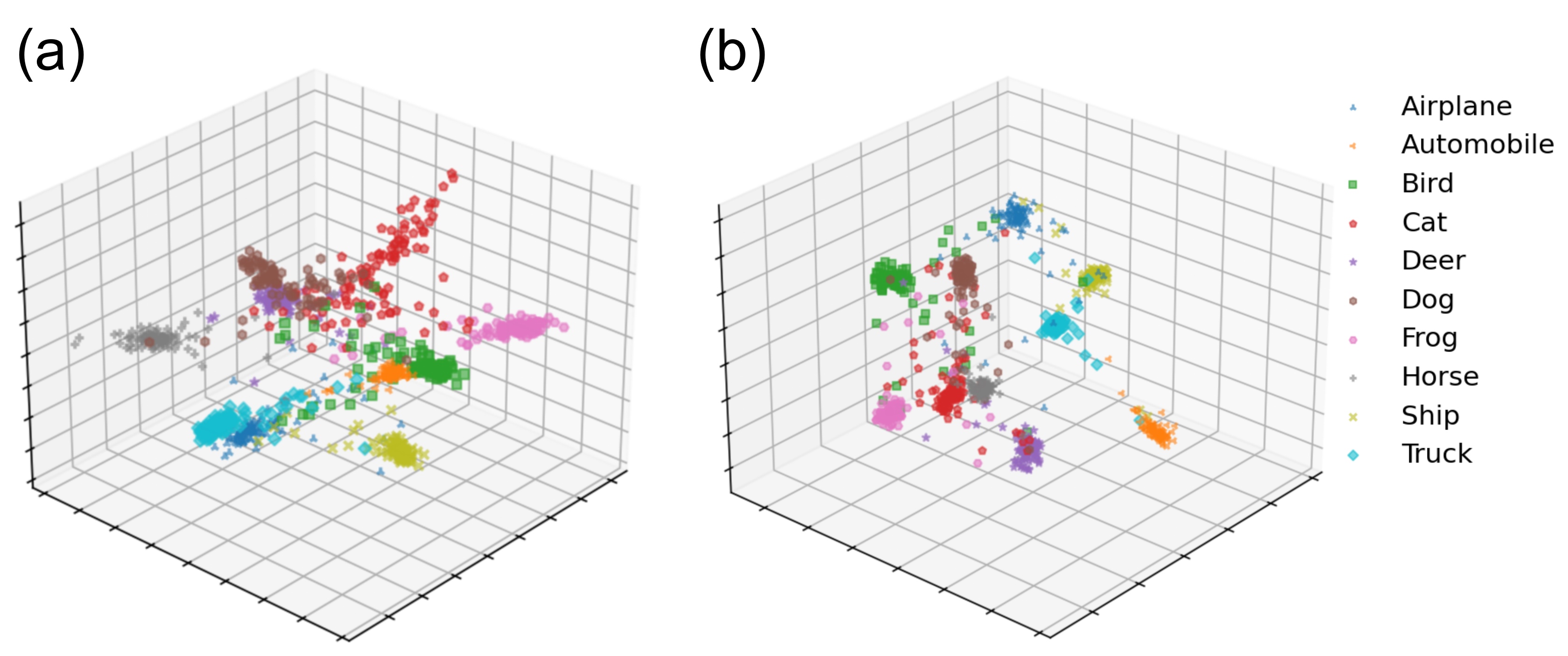}
      \caption{Latent space visualizations of the CvT models produced by (a) the baseline and (b) proposed method compared. }
      \label{fig:latent-vis}
   \end{figure}

   To better understand effects of the proposed method on model latent space, latent spaces of the baseline CvT and CvT model trained by applying the proposed CKFR method on the CIFAR-10 dataset were visualized. These visualizations were produced by adding a three-neuron fully-connected layer in front of the model classifier and plotting the three-dimensional latent distribution of this layer.
   
   As can be seen in Figure \ref{fig:latent-vis}, the proposed method produced a finer latent space with better categorical discriminativity and allocation. For example, the categories dog and cat share very similar backgrounds and surrounding objects and can be better distinguished in the model latent space optimized by the proposed CKFR, and transportation vehicle categories are better grouped together in the model latent space that applied the proposed method.

   \subsubsection{Evaluation of the $\alpha$ hyperparameter}
  
   \begin{figure}[htbp]
      \centering
      \begin{tikzpicture}
      \begin{axis}[small,grid=major,xmode=log,ymin=58.1,xlabel=$\alpha \text{ value}$,ylabel=$\text{Top-1 Acc. }(\%)$]
         \addplot[smooth,teal,mark=x]
         coordinates {
            (1.0,58.2) (3.0,59.0)
            (10.0,59.3) (30.0,59.6)}
            [every node/.style={yshift=6pt}]
               node [pos=0]    {$58.2$}
               node [pos=0.33]    {$59.0$}
               node [pos=0.66]    {$59.3$}
               node [pos=1.0]    {$59.6$}
         ;
         \addplot[violet,sharp plot,update limits=false]
            coordinates {(0.1,58.3) (100.0,58.3)}
            node[above] at (axis cs:5.0,58.3) {Baseline};
      \end{axis}
      \end{tikzpicture}
      \caption{Classification accuracies produced by applying varied $\alpha$ hyperparameter values. Baseline indicates baseline performance without applying the proposed method.}
      \label{fig:alphas}
   \end{figure}
   
   Figure \ref{fig:alphas} presents classification accuracies produced by applying the proposed method with the $\alpha$ hyperparameter value set to be 0.0, 1.0, 3.0, 10.0, and 30.0. The evaluations were performed using the ResNet-34 model trained on the CIFAR-100 dataset. Results demonstrate a strong positive correlation between the evaluated $\alpha$ values and produced classification accuracies, indicating effectiveness of the proposed method on enhancing image recognition performance.

   \subsubsection{Evaluation of the $\ell$ hyperparameter}

   \begin{figure}[htbp]
      \centering
      \includegraphics[width=0.9\linewidth]{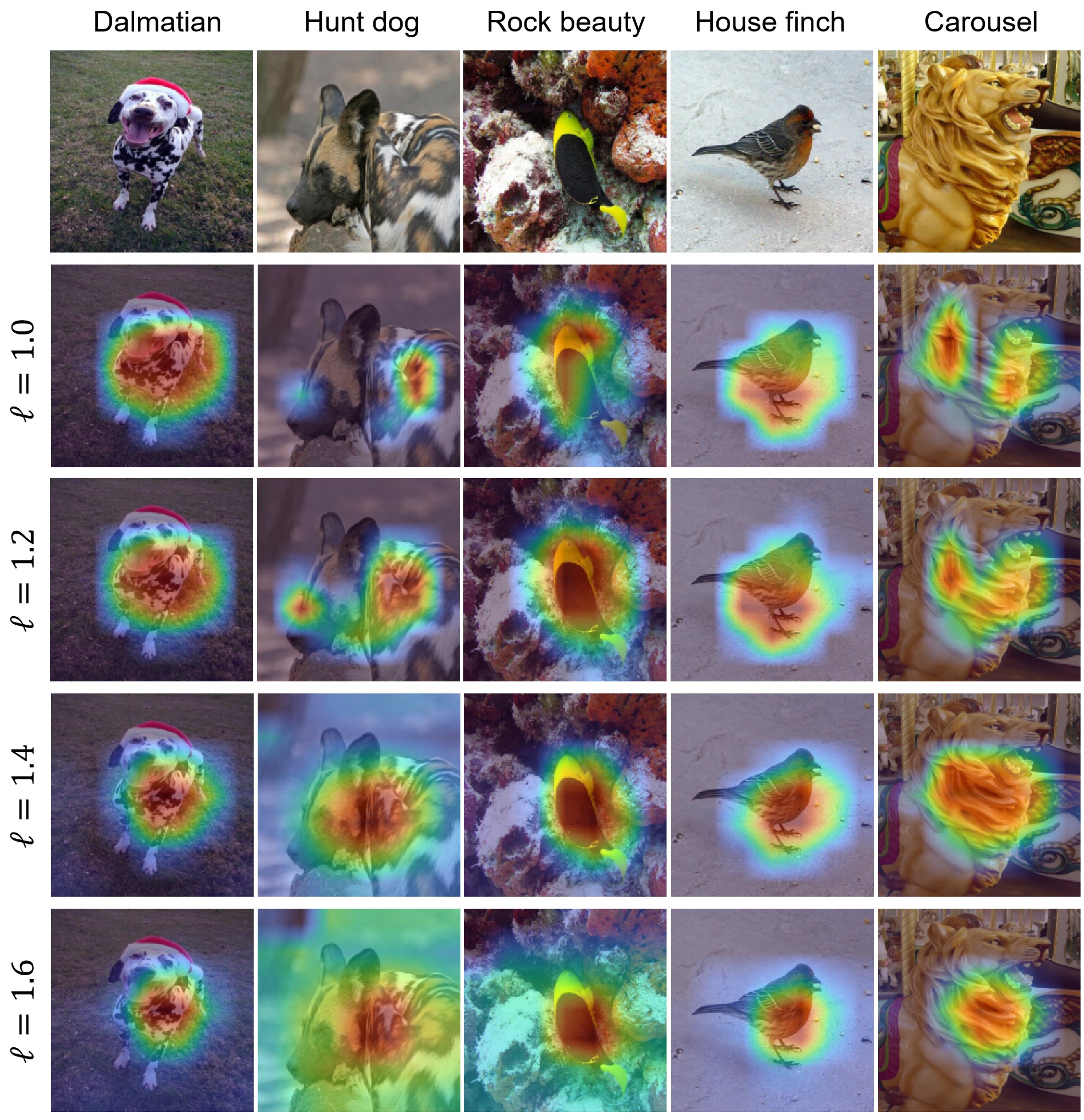}
      \caption{Grad-CAM visualizations generated by applying varied $\ell$ hyperparameter of the proposed method. }
      \label{fig:l-eval}
   \end{figure}

   \begin{figure*}[htbp]
      \centering
      \includegraphics[width=0.82\linewidth]{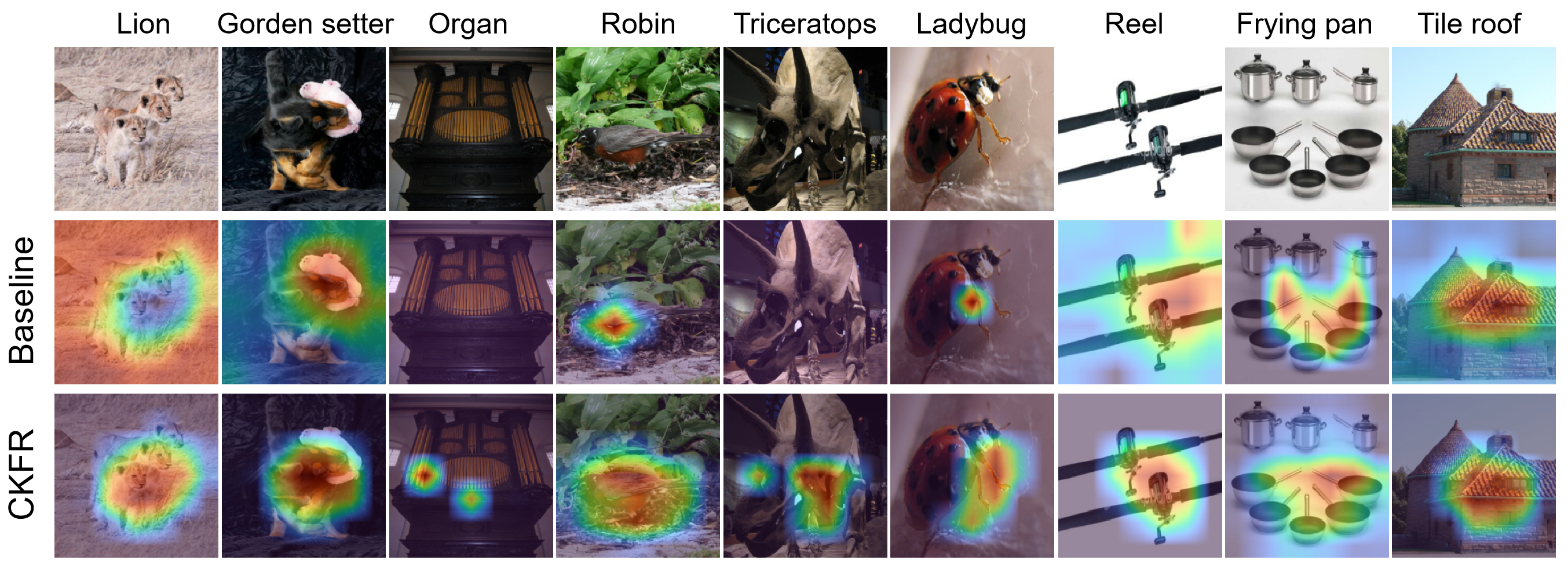}
      \caption{Grad-CAMs of the WResNet models produced from without (baseline) and with applying the proposed method (CKFR) compared. }
      \label{fig:gradcam2}
   \end{figure*}

   Figure \ref{fig:l-eval} illustrates the effects of varied $\ell$ hyperparameter of the proposed method ($\ell$ set to be 1.0, 1.2, 1.4, and 1.6) on the Grad-CAM visualization produced. It can be seen that a higher $\ell$ order value tends to produce disperse class activation and include more surrounding information in the prediction and be less specific.

   \subsubsection{CAM visualization}
   
   To examine the learned features, Gradient-weighted CAM (Grad-CAM) \cite{gradcam} visualizations of the WResNet models without (baseline) and with applying the proposed method were produced and demonstrated in Figure \ref{fig:gradcam2}. Compared to the baseline, the proposed method tremendously improved target object identification and localization performance and enhanced the reasoning aspect of image recognition.

   \section{Discussions}

   Image classification is a fundamental visual task in human recognition and can also be essential everyday-tools like lookup books that help to identify and understand image categories in real world. Hierarchical knowledge is like memory and contains class level similarity information about image categories. Usually, image classification model weights are updated in a supervised learning process with a focus on improving classification accuracy. In this work, we proposed an aligning method that fuses inter-class similarity knowledge about image categories with backbone image classification models in a weakly-supervised learning process. The proposed method optimizes model latent space to a structure that agrees with prior knowledge space and provides clues about the target object in the image. It helps to identify the target object in the processed image and enhance the reasoning aspect of the recognition.

   The proposed method works with image classification model with high latent dimensionality. It alters features captured by the image classification model compared to the baseline, i.e., image classification model trained by supervised learning only, and leads to more accurate focus on the target objects as demonstrated in Figure \ref{fig:gradcam2}. Quantitative evaluation results in terms of localization accuracies produced by applying the proposed method on the ImageNet-1K dataset also indicate that the proposed method tremendously improves the reasoning aspect of image classification. As a result, the proposed method can enhance robustness of image classification predictions. And hierarchical knowledge could be valuable prior to improve image recognition performances when fused by performing the proposed method.

   \section{Conclusion}
   
   We have proposed a novel method based on aligning and deep metric learning to fuse hierarchical knowledge about image categories with image classification backbone models and to improve image recognition performance, especially the reasoning aspect of the recognition. The proposed method incorporates a new triplet loss function term that is simple yet effective in aligning distance in the deep neural network model latent space with distance in the knowledge space in an end-to-end manner. Experimental results indicate the effectiveness of the proposed method in improving image recognition performance including the classification accuracy evaluated on the CIFAR-10 and CIFAR-100 datasets, as well as the reasoning aspect of the recognition in terms of WSOL performance evaluated on the Mini-ImageNet and ImageNet1K datasets.

   \vspace{20pt}

   \bibliographystyle{IEEEtran}
   \bibliography{IEEEabrv,ref.bib}

\end{document}